\documentclass[letterpaper, 10 pt, conference]{ieeeconf}  

\IEEEoverridecommandlockouts                              

\usepackage{amsmath}
\usepackage{amssymb}
\usepackage{booktabs}
\usepackage{graphicx}
\usepackage{url}

\title{\LARGE \bf
	Manipulation-Feasible Navigation Among Movable Obstacles with Discrete-Contact Pushing
}

	\author{Shaohu Wang, Aiguo Song$^{\dagger}$, Yulong Yuan, Zhongyu Sun, Tianyuan Miao and Qinjie Ji
		\thanks{Shaohu Wang, Aiguo Song, Yulong Yuan, Zhongyu Sun, Tianyuan Miao and Qinjie Ji are with the State Key Laboratory of Bioelectronics, Jiangsu Key Laboratory of Robot Perception and Control Technology, School of Instrument Science and Engineering, Southeast University, Nanjing 210096, China (e-mail: wangsh@seu.edu.cn, a.g.song@seu.edu.cn). \textit{(Corresponding author: Aiguo Song)}}
	}

\begin{document}

		\maketitle
		\thispagestyle{empty}
		\pagestyle{empty}

		
		\begin{abstract}
			
			In environments with large movable obstacles, detour-only navigation can be inefficient or even infeasible, while obstacle interaction requires reasoning about navigation benefit, feasible placement, and executable manipulation. We present a hierarchical navigation among movable obstacles (NAMO) framework for mobile manipulators. At the high level, the planner identifies key blocking obstacles from reference paths and searches for relocation plans that jointly satisfy geometric, manipulation, and downstream navigation constraints. When direct relocation is hindered by other movable objects, a large language model (LLM) is selectively invoked to infer auxiliary manipulation dependencies, which are then verified by deterministic geometric planning. To execute the resulting relocation goals, we define discrete contact modes on the surfaces of box-shaped obstacles and select contact faces and regions online based on position and orientation errors, enabling straight, side, and corner pushing through contact switching. A recurrent reinforcement-learning policy coordinates the mobile base and manipulator to track tool center point (TCP) targets while preserving end-effector reachability during sustained pushing. Simulation and real-robot experiments demonstrate feasible navigation-manipulation in detour, single- and multi-obstacle relocation, and dependency-constrained scenarios, validating the framework for interactive navigation with large non-graspable obstacles. The open-source project is available at \url{https://cloudytosunny.github.io/NAMO_DCPushing/}.
			
			
		\end{abstract}

		\section{INTRODUCTION}
		
		Mobile robots have been widely deployed in applications such as warehouse logistics, inspection in hazardous or hard-to-access environments, and public safety. Conventional navigation systems, however, typically treat surrounding objects as non-traversable obstacles, and detouring around movable objects can substantially increase navigation cost or even render the goal unreachable \cite{zhangNavigationMovableObstacles2023}. Navigation among movable obstacles addresses this limitation by allowing robots to actively interact with movable obstacles to recover or improve navigable space, thereby enabling more efficient navigation in constrained environments  \cite{renSearchBasedPathPlanning2025,weedaPushingClutterMovability2025}.

\begin{figure}[t]
			\centering
			\includegraphics[scale=0.28]{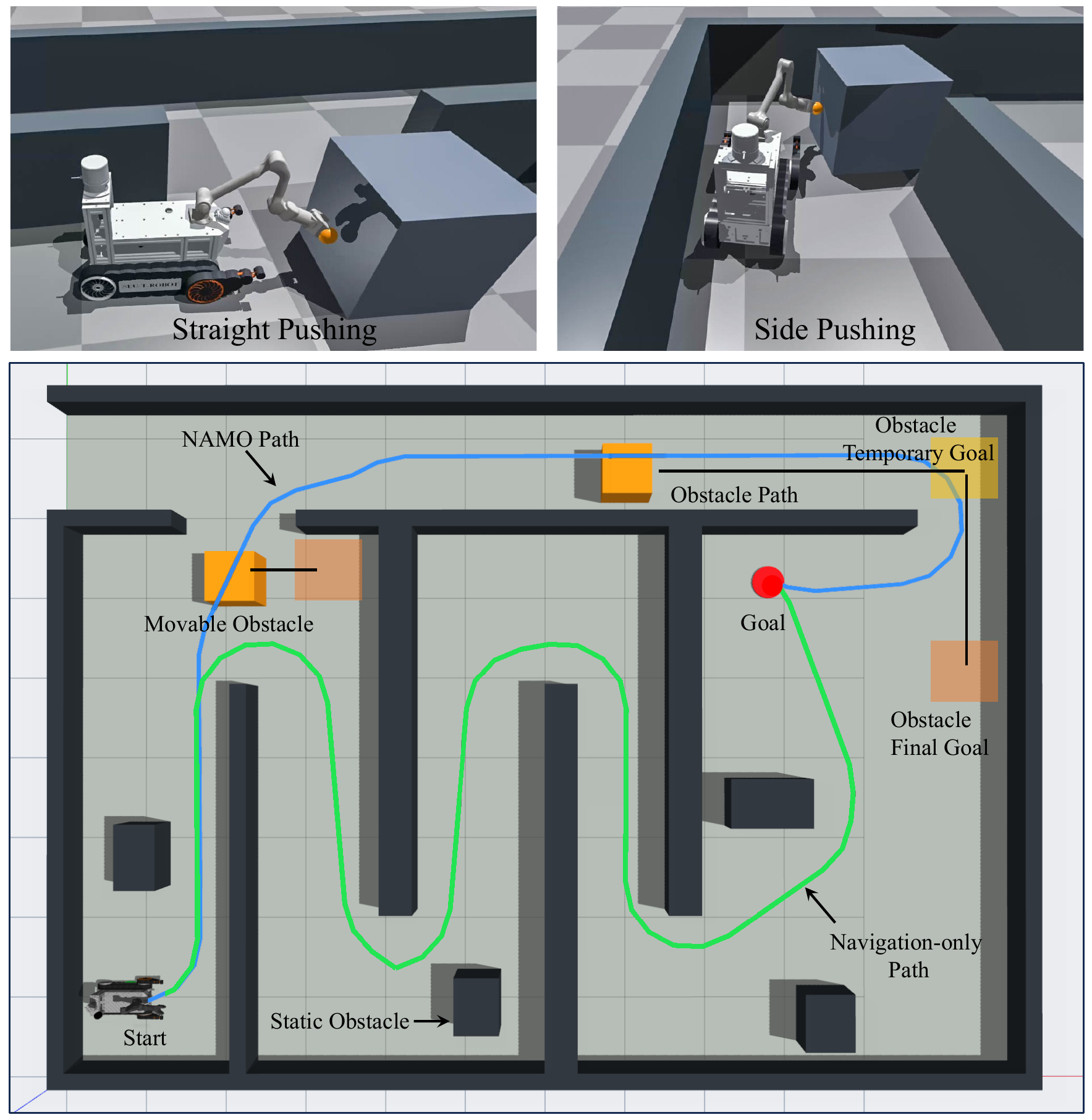}
			\caption{Illustration of manipulation-feasible NAMO with discrete contact modes. Straight and side pushing skills are composed from the discrete contact set. The green curve denotes the navigation-only route, while the blue curve denotes the NAMO route, where the obstacle is pushed to a planned placement to restore traversability. The proposed framework selects the lower-cost route for efficient goal reaching.}
			\label{f1}
		\end{figure}

		Small movable objects can often be removed through grasping, whereas large box-shaped obstacles may instead require nonprehensile pushing to clear the way \cite{tangUnwieldyObjectDelivery2023,dadiotisDynamicObjectGoal2025}. Such NAMO tasks therefore couple navigation with obstacle relocation: when multiple routes are available, the robot must decide whether to detour or interact, identify the obstacles that truly constrain the candidate routes, and determine relocation poses that improve accessibility while remaining physically executable, as illustrated in Fig.~\ref{f1} \cite{EfficientNavigationMovable2025b}. Moreover, route-blocking obstacles may not be the only objects that must be moved, since other movable objects can occupy the required placement, approach, or pushing space, creating inter-object manipulation dependencies \cite{ahnReloPushBOSSOptimizationGuidedNonmonotone2026}. Although recent work has explored push-based multi-object relocation \cite{ahnReloPushMultiObjectRearrangement2025} and LLM-assisted NAMO reasoning \cite{zhangNAMOLLMEfficientNavigation2025}, jointly reasoning about navigation benefit, feasible obstacle relocation, and mobile-manipulation constraints remains challenging.
		

		Even when a high-level planner finds a collision-free relocation path and target pose, the motion may still be physically infeasible for a mobile manipulator because the obstacle itself is unactuated and must be moved through robot contact. For large box-shaped objects, pushing behavior is strongly contact-dependent: near-central contacts mainly induce translation, whereas offset contacts introduce moments that couple translation and rotation, making contact switching, base repositioning, and re-contact necessary when the desired motion changes direction \cite{zhangSwitchingPushingSkill2023}. Reinforcement learning has shown promise for coordinating the mobile base and manipulator to push large objects toward target poses \cite{dadiotisDynamicObjectGoal2025}, and has also been incorporated into mobile-manipulator NAMO for whole-body pushing execution \cite{EfficientNavigationMovable2025b}. However, complex relocation cannot be effectively handled by separately trained pushing skills. This motivates a unified interface that maps target object poses to discrete contact modes and continuous whole-body control, allowing a finite set of contact modes to be composed into flexible multi-stage pushing behaviors.
		
\begin{figure*}[t]
				\setlength{\belowcaptionskip}{-0.5cm}
				\vspace{0.2cm}
				\centering
				\includegraphics[scale=0.38]{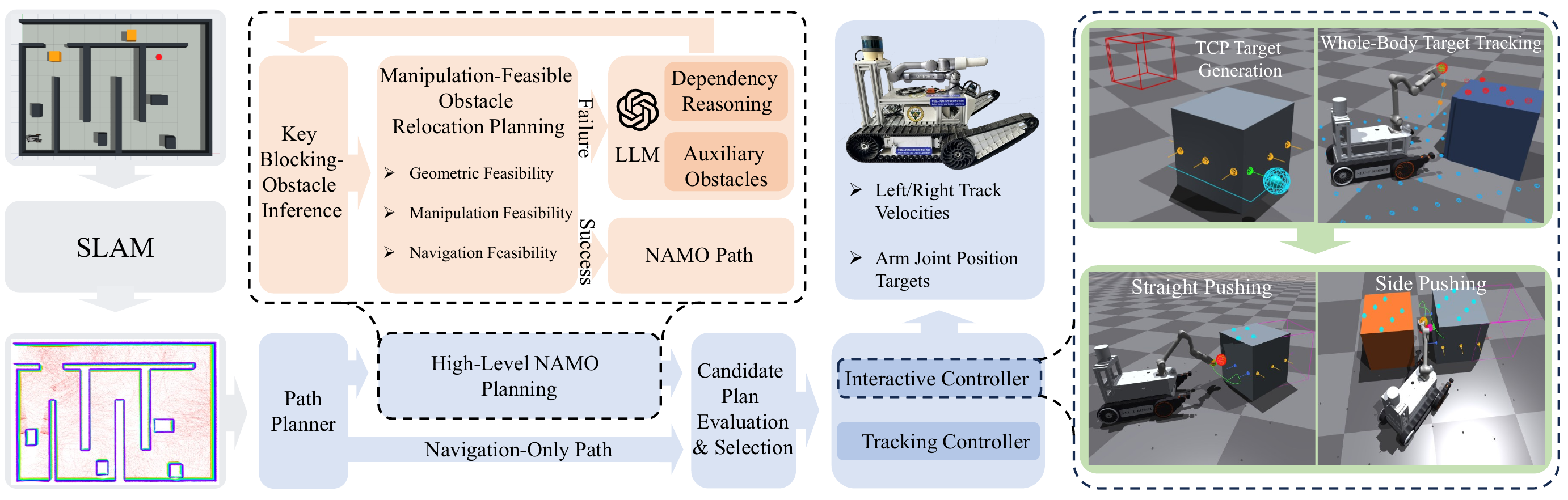}
				\caption{Overview of the proposed hierarchical NAMO framework. The planning layer evaluates navigation-only and NAMO hypotheses and selects the feasible plan with the lowest total cost. NAMO planning identifies key blockers and searches for manipulation-feasible relocation poses; when direct relocation fails, an LLM is selectively invoked to infer auxiliary obstacles and their manipulation dependencies for replanning. The execution layer maps relocation goals to discrete contact modes and TCP reference targets, which are tracked by a learned whole-body controller to compose straight, side, and corner pushing, while navigation-only segments follow the selected collision-free path.}
				\label{f2}
			\end{figure*}

		
		To address these challenges, we consider NAMO for mobile manipulators in known environments with large, non-graspable but pushable box-shaped obstacles, and propose a hierarchical navigation--manipulation framework shown in Fig.~\ref{f2}. At the high level, the planner evaluates navigation-only and interaction-enabled hypotheses, identifies key blockers from reference paths, and searches for relocation poses that satisfy geometric, manipulation, and downstream reachability constraints. When direct relocation fails because of insufficient manipulation space, an LLM is selectively invoked to infer auxiliary obstacles and their manipulation dependencies, while all inferred hypotheses are subsequently verified by deterministic planning. At the execution level, target object poses are mapped online to discrete contact modes and corresponding TCP references, which are tracked by a reinforcement-learning-based whole-body controller coordinating the mobile base and manipulator. Contact-mode switching further enables straight, side, and corner pushing to be composed within a unified control framework. This hierarchical design provides a direct interface between task-level obstacle relocation and physically executable mobile manipulation.
		
		
		The main contributions of this work are summarized as follows:

			\begin{itemize}
				\item \emph{A manipulation-feasible high-level NAMO planning method for large, non-graspable box-shaped obstacles.} It jointly evaluates navigation and interaction options, identifies key blockers, and searches for executable relocation poses, with LLM-assisted dependency reasoning when direct relocation fails.
				
				\item \emph{A discrete-contact mobile-manipulation method with learned whole-body control.} Target object poses are mapped to discrete contact modes and TCP references, enabling a shared policy to realize straight, side, and corner pushing through contact switching.
				
				\item \emph{An integrated mobile-manipulator NAMO system validated in simulation and real-world environments.} Experiments demonstrate effective decision making, obstacle relocation, multi-contact pushing, and integrated planning and execution in constrained and dependency-aware scenarios.
			\end{itemize}

			\section{Related Work}
			
			\subsection{Navigation Among Movable Obstacles}
			
			NAMO extends conventional navigation by allowing robots to actively modify the environment when interaction provides a feasible or more efficient route. Existing work has studied task-level trade-offs between detouring and manipulation \cite{zhangNavigationMovableObstacles2023}, interaction under partial or unknown environments \cite{muguiraiturraldeVisibilityAwareNavigationMovable2023,InteractiveFARInteractiveFast2024}, and perception-driven interactive navigation \cite{schochINSightInteractiveNavigation2024}. More recent methods couple robot and object motion more tightly through joint-state search or movability-aware interaction planning \cite{renSearchBasedPathPlanning2025,weedaPushingClutterMovability2025}, while multi-robot NAMO further exposes conflicts and deadlocks induced by shared movable obstacles \cite{renaultMultiRobotNavigationMovable2024}.
			
			For multi-object tasks, ReloPush and ReloPush-BOSS address sequential and nonmonotone obstacle relocation \cite{ahnReloPushMultiObjectRearrangement2025,ahnReloPushBOSSOptimizationGuidedNonmonotone2026}, whereas NAMO-LLM uses language-model guidance to bias obstacle selection, placement, and ordering \cite{zhangNAMOLLMEfficientNavigation2025}. In contrast, our method couples route-conditioned blocker identification with manipulation-feasible placement and failure-driven dependency expansion. The LLM is invoked only after deterministic relocation planning fails, and all inferred dependencies are subsequently verified by geometric and manipulation constraints.
			
			\subsection{Nonprehensile Manipulation for Obstacle Relocation}
			
			Nonprehensile pushing enables robots to relocate large objects without grasping, but requires coupling desired object motion with executable contact and robot motion. Model-based approaches incorporate contact and motion constraints into pushing control \cite{tangUnwieldyObjectDelivery2023}, while recent work coordinates planning and control for object pushing \cite{bertoncelliStreamliningObjectPushing2024} or learns pushing behaviors under uncertain object properties \cite{leeGoalDrivenRoboticPushing2025,dadiotisDynamicObjectGoal2025}. Object-centric planning and learned mobile manipulation further connect desired object motion with closed-loop pushing execution \cite{renObjectCentricKinodynamicPlanning2025,EfficientNavigationMovable2025b}.
			
			Contact selection is particularly important when both translation and rotation are required. Switching Pushing Skill combines discrete pushing-point decisions with continuous feedback control \cite{zhangSwitchingPushingSkill2023}, while learned arm-pushing has been applied to large-obstacle relocation during interactive navigation \cite{biInteractiveNavigationLegged}. In contrast, our method uses discrete contact modes as an explicit interface between object-level relocation goals and whole-body execution, mapping target object poses to TCP references and composing straight, side, and corner pushing through contact switching with a shared learned controller.

			%
			%
			%
			%
			%
			%

			\section{Methodology}

			\subsection{Problem Formulation}
			We consider a mobile-manipulator NAMO problem in a known static environment containing fixed and movable obstacles, represented as
			\begin{equation}
				\mathcal{E}
				=
				\left\{
				\mathcal{O}_{s},
				\mathcal{O}_{m},
				\mathcal{F}
				\right\},
				\label{eq:environment}
			\end{equation}
			where $\mathcal{O}_{s}$ and $\mathcal{O}_{m}$ denote the sets of fixed and movable box-shaped obstacles, respectively, and $\mathcal{F}$ is the collision-free workspace. The robot state is $q_{r} = \left(q_{b},q_{a}\right)$, with $q_{b}=(x_{b},y_{b},\theta_{b})$ denoting the planar base pose and $q_{a}\in\mathbb{R}^{6}$ denoting the manipulator configuration. Each movable obstacle $o_i\in\mathcal{O}_{m}$ has a planar pose $q_i=(x_i,y_i,\theta_i)$.
			
			We assume that the map, obstacle movability labels, initial poses, and geometric dimensions are known. Movable obstacles are non-graspable but pushable on the ground plane, and navigation collision checking uses the full planar robot footprint.
			Given an initial robot state \(q_r^{s}\) and a navigation goal \(q_r^{g}\), the planner searches over task hypotheses
			\begin{equation}
				\mathcal{H}
				=
				\left(
				\pi,
				\mathcal{B},
				\sigma,
				\mathcal{G}
				\right),
				\mathcal{G}=\{g_i \mid o_i\in\mathcal{B}\},
				\label{eq:task_hypothesis}
			\end{equation}
			where $\pi$ is the navigation route, $\mathcal{B}\subseteq\mathcal{O}_{m}$ is the set of obstacles selected for relocation, $\sigma$ specifies their manipulation order, and $\mathcal{G}$ contains the corresponding target poses. When $\mathcal{B}=\varnothing$, the hypothesis becomes navigation-only.
			
			Among all feasible hypotheses, the planner selects
			\begin{equation}
				\mathcal{H}^{*}
				=
				\arg\min_{\mathcal{H}\in\mathbb{H}_{f}}
				\left[
				w_{n}J_{\mathrm{nav}}(\mathcal{H})
				+
				w_{m}J_{\mathrm{man}}(\mathcal{H})
				\right],
				\label{eq:task_objective}
			\end{equation}
			where $\mathbb{H}_{f}$ denotes the feasible hypothesis set, $J_{\mathrm{nav}}$ the accumulated navigation cost measured by the planned path length, and $J_{\mathrm{man}}$ the manipulation cost, including obstacle displacement, robot repositioning, and contact switching. The weights $w_n$ and $w_m$ balance navigation efficiency and manipulation effort.

			\subsection{High-Level NAMO Decision and Hierarchical Obstacle Relocation Planning}
			
			\subsubsection{Path-Conditioned Key Blocker Inference}
			We first use Hybrid A* \cite{dolgovPracticalSearchTechniques2008} to plan from $q_r^{s}$ to $q_r^{g}$ in the complete environment. If a feasible path exists, it is retained as a navigation-only candidate $\pi^{\mathrm{nav}}$. We then temporarily remove all movable obstacles and compute a set of reference paths $\pi^{\mathrm{ref}}_k$ to identify corridors that may benefit from obstacle interaction.
			
			The robot footprint with a safety margin is swept along $\pi^{\mathrm{ref}}_k$ to form the traversability corridor $\mathcal{C}(\pi^{\mathrm{ref}}_k)$. Movable obstacles intersecting this corridor form the initial blocking set
			\begin{equation}
				\widetilde{\mathcal{B}}_{k}
				=
				\left\{
				o_i\in\mathcal{O}_{m}
				\;\middle|\;
				\Omega(o_i)\cap\mathcal{C}(\pi^{\mathrm{ref}}_k)
				\neq\varnothing
				\right\},
				\label{eq:initial_blockers}
			\end{equation}
			where $\Omega(o_i)$ denotes the planar region occupied by obstacle $o_i$. Since corridor intersection alone does not imply a necessary relocation, each candidate obstacle is restored individually and the path is replanned. The key blocker set is
			\begin{equation}
				\mathcal{B}_{k}
				=
				\left\{
				o_i\in\widetilde{\mathcal{B}}_{k}
				\;\middle|\;
				\operatorname{Plan}
				\left(
				\mathcal{E}_{k}^{(i)},
				q_r^{s},
				q_r^{g}
				\right)
				=
				\varnothing
				\right\},
				\label{eq:key_blockers}
			\end{equation}
			where $\mathcal{E}_{k}^{(i)}$ is the reference environment with $o_i$ restored. Each resulting pair $(\pi^{\mathrm{ref}}_k,\mathcal{B}_k)$ is then passed to manipulation-feasible obstacle relocation planning and task-level evaluation.

			\subsubsection{Manipulation-Feasible Obstacle Relocation Planning}
			\label{M2.2.2}
			
			For each key blocker $o_i$, candidate target poses $g_i=(x_i^g,y_i^g,\theta_i^g)$ are sampled hierarchically in progressively expanded neighborhoods around its current pose. Nearby collision-free placements are considered first to avoid unnecessary obstacle displacement.
			For each sampled target $g_i$, an obstacle relocation path
			\begin{equation}
				\pi_i^o(g_i)=
				\{q_i^0,q_i^1,\ldots,q_i^K\},
				q_i^0=q_i,\quad q_i^K=g_i,
				\label{eq:object_push_path}
			\end{equation}
			is searched in the planar object state space $(x_i,y_i,\theta_i)$. Each transition $q_i^k\!\rightarrow q_i^{k+1}$ is generated by an admissible pushing mode $m_k$, while collision states are rejected during search.
			For each collision-free object path, downstream navigation is verified by
			replanning in the updated environment:
			\begin{equation}
				\Phi_{\mathrm{nav}}(g_i)
				=
				\mathbb{I}\!\left[
				\exists\,\pi_i^r:
				q_{r,i}^{+}\leadsto q_{r,i}^{\mathrm{next}},
				\;
				\pi_i^r
				\subset
				\mathcal{F}\!\left(\mathcal{E}_i^{+}(g_i)\right)
				\right].
				\label{eq:downstream_navigation}
			\end{equation}
			where $q_{r,i}^{+}$ is the post-relocation robot state, and
			$q_{r,i}^{\mathrm{next}}$ denotes the pre-manipulation state of the next
			blocker or the final navigation goal. $\mathcal{E}_i^{+}(g_i)$ denotes the updated environment after
			relocating $o_i$ to $g_i$.
			
			The remaining candidates are further checked for manipulation
			executability:
			\begin{equation}
				\Phi_{\mathrm{man}}(\pi_i^o)
				=
				\mathbb{I}\left[
				\begin{aligned}
					&\mathcal{S}(q_i^k,m_k)\subseteq\mathcal{F}_r, k=0{:}K-1\\
					&\Gamma(q_i^k,m_k,q_i^{k+1})=1,
					k=0{:}K-1
				\end{aligned}
				\right],
				\label{eq:manipulation_feasibility}
			\end{equation}
			where $\mathcal{S}(q_i^k,m_k)$ denotes the robot configurations required
			by mode $m_k$, $\mathcal{F}_r$ is the collision-free reachable robot
			space, and $\Gamma$ evaluates whether the corresponding object transition,
			including contact switching when required, is executable.
			
			Candidates satisfying both navigation $\Phi_{\mathrm{nav}}=1$ and manipulation feasibility $\Phi_{\mathrm{man}}=1$ are
			ranked by
			\begin{equation}
				J_{\mathrm{man}}(\pi_i^o)
				=
				\alpha L_o(\pi_i^o)
				+
				\beta N_{\mathrm{sw}}(\pi_i^o)
				+
				\gamma L_r(\pi_i^o),
				\label{eq:placement_cost}
			\end{equation}
			where $L_o$ is the obstacle travel distance, $N_{\mathrm{sw}}$ is the
			number of contact-mode switches, and $L_r$ is the robot approach and
			repositioning cost. If no feasible candidate is found locally, the sampling
			region is progressively expanded; for multiple blockers, the environment
			is updated after each accepted relocation and backtracked when necessary.

			\subsubsection{Failure-Driven Manipulation Dependency Expansion}
			
			The blocking set $\mathcal{B}_k$ contains obstacles that directly affect the candidate route, yet some blockers may not be directly relocatable. For a blocker $o_i$, if no candidate relocation satisfies the geometric, downstream-navigation, and manipulation-feasibility constraints, direct relocation fails. This may occur because other movable objects occupy the required placement, approach, or pushing space. We therefore selectively invoke an LLM using structured failure context:
			\begin{equation}
				\left(
				\mathcal{A}_i,
				\mathcal{D}_i
				\right)
				=
				\operatorname{LLM}
				\left(
				\mathcal{X}_i^{\mathrm{fail}}
				\right),
				\mathcal{D}_i
				=
				\left\{
				o_j\prec o_i
				\mid
				o_j\in\mathcal{A}_i
				\right\},
				\label{eq:dependency_inference}
			\end{equation}
			where $\mathcal{X}_i^{\mathrm{fail}}$ contains the identified key blockers, local map information, task description, and failure reason. The output $\mathcal{A}_i$ denotes the inferred auxiliary obstacles, while $\mathcal{D}_i$ specifies their manipulation dependencies; $o_j\prec o_i$ indicates that $o_j$ should be relocated before $o_i$.
			
			The LLM does not generate continuous object poses or relocation paths. Instead, each inferred auxiliary obstacle is returned to the deterministic relocation planner in Sec.~\ref{M2.2.2} for feasibility verification. Feasible relocations update the environment before retrying the original blocker; otherwise, the corresponding hypothesis is rejected.

			\begin{figure}[t]
				\centering
				\includegraphics[scale=0.18]{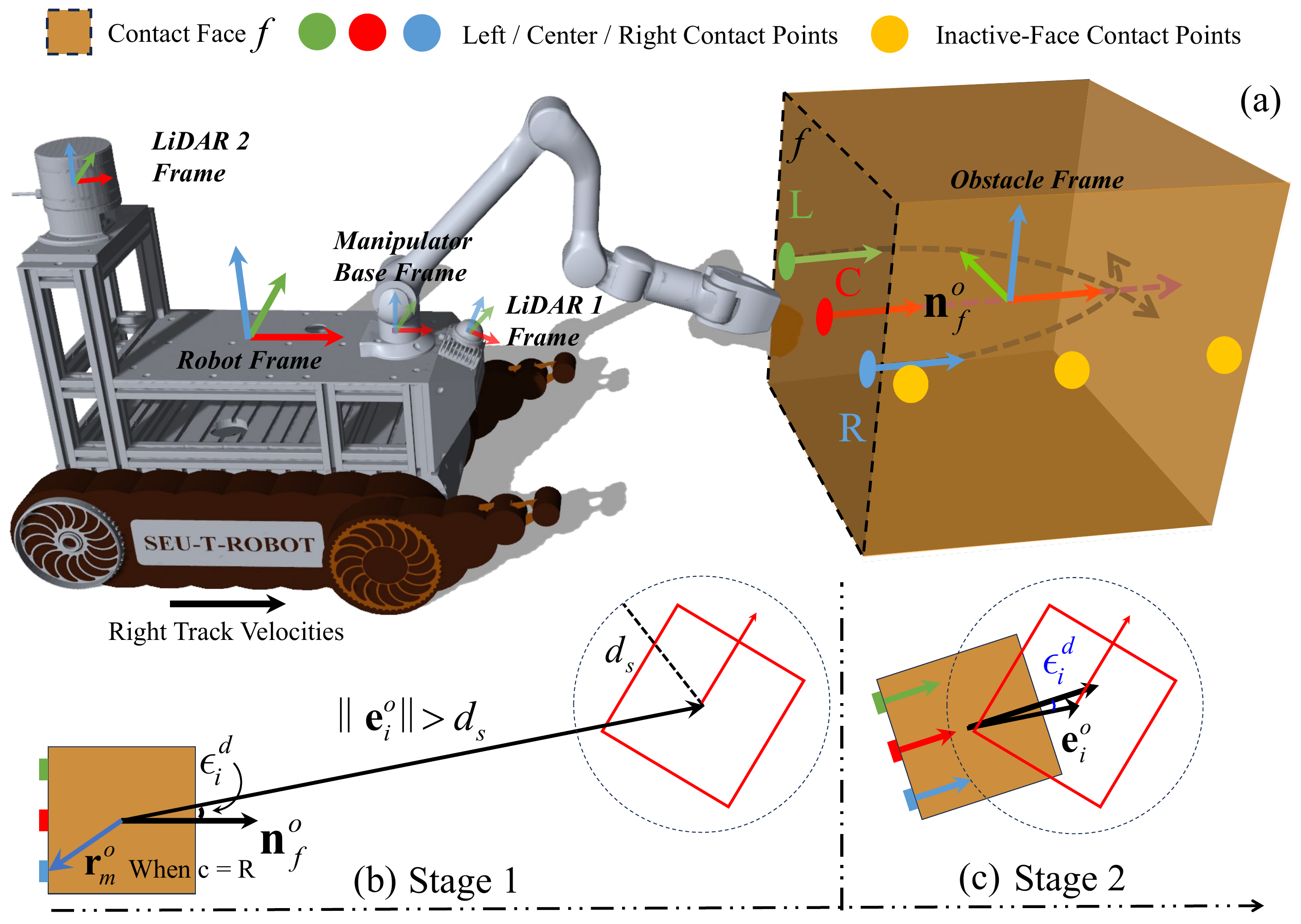}
				\caption{Robot--obstacle interaction with discrete contact modes. (a) Each vertical box face contains left, center, and right contact anchors, which define TCP reference targets and induce different translational and rotational effects. (b) In Stage~1, the pushing direction is aligned with the target before straight pushing. (c) In Stage~2, contact selection jointly considers position and orientation errors to reach the target pose.}
				\label{f3}
			\end{figure}

			\subsection{Learned Mobile Manipulation via Discrete Contact Modes}
			
			Given the target pose $g_i$ produced by the high-level planner, we map the
			object-level relocation goal to a discrete contact mode and a corresponding
			TCP reference target. A learned whole-body policy then coordinates the mobile
			base and manipulator to track this target during pushing.
			
			
			\subsubsection{Discrete Contact Modes on Box Surfaces}
			
			\label{D2.3.1}
			
			As illustrated in Fig.~\ref{f3}(a), each vertical face of obstacle $o_i$ is partitioned into left, center, and right
			contact regions, represented by fixed anchors in the object frame. A contact mode is defined as
			\begin{equation}
				m=(f,c),\qquad
				f\in\mathcal{T}_i,\quad c\in\{L,C,R\},
				\label{eq:contact_mode}
			\end{equation}
			where $\mathcal{T}_i$ contains the four vertical faces. For each face,
			$\mathbf n_f^o$ is the unit inward normal from the face center to the object
			center and represents its nominal pushing direction.
			
			Given $q_i=(x_i,y_i,\theta_i)$ and $g_i=(x_i^g,y_i^g,\theta_i^g)$, the target
			displacement in the object frame and the selected contact face are
			\begin{equation}
				\mathbf e_i^o
				=
				R^\top(\theta_i)
				\begin{bmatrix}
					x_i^g-x_i\\ y_i^g-y_i
				\end{bmatrix},
				f_i^*
				=
				\arg\max_{f\in\mathcal T_i}
				\frac{(\mathbf n_f^o)^\top\mathbf e_i^o}
				{\|\mathbf e_i^o\|}.
				\label{eq:contact_face_selection}
			\end{equation}
			
			For each candidate contact mode $m$, its
			nominal short-push effect is geometrically approximated by
			\begin{equation}
				\Delta\mathbf p_m^o=s_m\mathbf n_f^o,
				\Delta\theta_m=
				\kappa_m(\mathbf r_m^o\times\mathbf n_f^o)_z,
				\label{eq:nominal_contact_motion}
			\end{equation}
			where $\mathbf r_m^o$ points from the object center to the contact anchor,
			$s_m$ is the nominal short-push displacement, and $\kappa_m$ scales the
			corresponding rotational effect. These increments characterize the local
			motion tendency of each candidate mode and are used only for contact-mode
			selection.
			
			
			The pushing process toward the target pose is divided into two stages.
			When $\|\mathbf e_i^o\|>d_s$, Stage~1 prioritizes motion toward the target, as illustrated in Fig.~\ref{f3}(b).
			Let $\epsilon_i^d$ denote the signed angle between
			$\mathbf n_{f_i^*}^o$ and $\mathbf e_i^o$. If
			$|\epsilon_i^d|\leq\epsilon_s$, only the position term is considered and
			the center contact is preferred for straight pushing. Otherwise, the
			position term is ignored and the orientation term is used to align the
			current pushing direction with the target direction. When
			$\|\mathbf e_i^o\|\leq d_s$, Stage~2 jointly considers the target position
			and final orientation, as illustrated in Fig.~\ref{f3}(c). The contact mode is selected by
			\begin{equation}
				\begin{aligned}
					m_i^*=\arg\min_{m\in\mathcal M_i}\Big[
					&\lambda_p^{s}
					\|\mathbf e_i^o-\Delta\mathbf p_m^o\|^2\\
					&+\lambda_\theta^{s}
					|\operatorname{wrap}
					(\Delta\theta_i^{\rm ref}-\Delta\theta_m)|^2
					\Big],
				\end{aligned}
				\label{eq:contact_selection}
			\end{equation}
			where $s\in\{1,2\}$ denotes the current stage, $\mathcal M_i$ is the candidate contact-mode set on the selected face, and $\lambda_p^s$ and \(\lambda_\theta^s\) activate the corresponding terms. $\Delta\theta_i^{\rm ref}$ denotes the direction-alignment error in Stage~1 or the final orientation error in Stage~2. When the translational error is negligible, the current face is retained for orientation regulation.
			During pushing, the current object pose and target distance are updated online; the nominal short-push effects of the candidate modes are recomputed accordingly, and Eq.~\eqref{eq:contact_selection} is reevaluated to select the current contact mode.
			The selected contact anchor is transformed into the world frame to define the TCP reference target for the whole-body controller.

			\begin{figure}[t]
				\centering
				\includegraphics[scale=0.17]{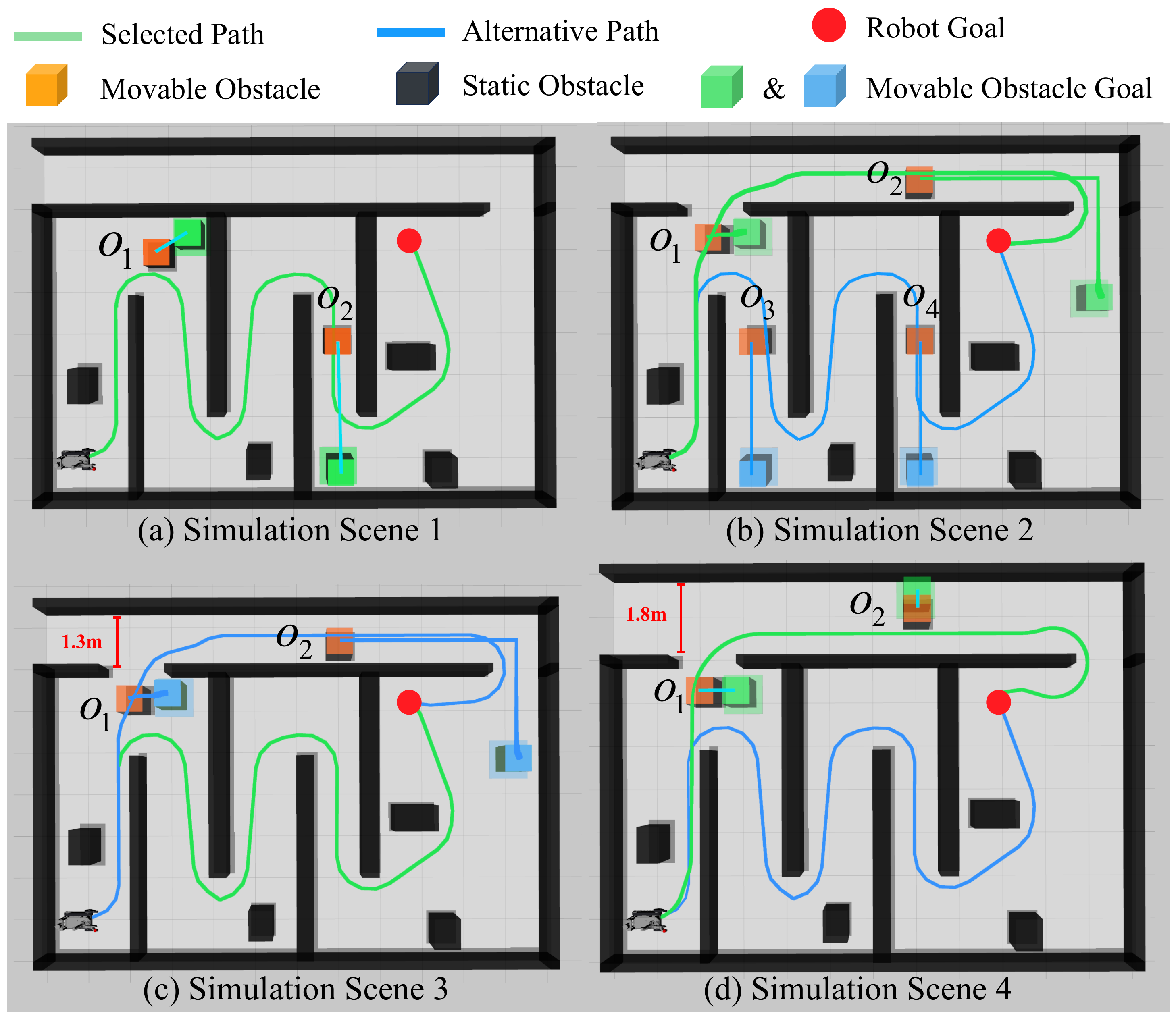}
				\caption{Planning results in four representative scenarios. Green denotes the selected route and obstacle relocation poses. (a) and (b) contain no feasible navigation-only path, whereas (c) and (d) show different route selections caused by small environmental changes.}
				\label{f4}
			\end{figure}
			
			\subsubsection{Reinforcement-Learning-Based Whole-Body TCP Target Tracking}
			
			During pushing, the TCP reference evolves with the obstacle motion, and
			manipulator-only tracking may lead to workspace limitations. We therefore
			train a recurrent whole-body policy using PPO
			\cite{schulmanProximalPolicyOptimization2017} to coordinate the mobile base
			and manipulator:
			\begin{equation}
				\begin{aligned}
					h_t &= \operatorname{GRU}(\mathbf z_t,h_{t-1}),\\
					[a_t^{L},a_t^{R},\Delta\mathbf p_t^{e}]
					&=\pi_\theta(\mathbf z_t,h_t),
				\end{aligned}
				\label{eq:whole_body_policy}
			\end{equation}
			where $\mathbf z_t$ denotes the policy observation, comprising the base
			orientation and planar motion, track velocities, arm joint positions and
			velocities, current TCP tracking error and target velocity, short-horizon
			TCP target previews, and a local occupancy grid describing the surrounding
			obstacles. $a_t^{L}$ and $a_t^{R}$ are the left- and right-track commands,
			and $\Delta\mathbf p_t^{e}\in\mathbb R^3$ is a Cartesian TCP residual
			applied to the reference target generated in Sec.~\ref{D2.3.1}. The corrected TCP
			target is converted to manipulator joint targets through damped
			least-squares inverse kinematics.
			
			Base motion and TCP tracking are optimized with separate value functions
			and reward objectives, accounting for workspace reachability, target
			tracking, collision avoidance, motion smoothness, and control effort. Their
			advantages are weakly coupled during PPO optimization:
			\begin{equation}
				\widetilde A_b=A_b+\rho A_e,
				\qquad
				\widetilde A_e=A_e+\rho A_b,
				\label{eq:cross_advantage}
			\end{equation}
			where $A_b$ and $A_e$ denote the base and end-effector advantages,
			respectively, and $\rho$ controls their coupling. This encourages base
			repositioning to support TCP tracking while allowing the manipulator to
			adapt to base motion.
			
			\subsubsection{Contact Switching and Pushing Skill Composition}
			
			
			During obstacle relocation, the desired pushing direction may change across
			different motion segments, requiring transitions between contact faces beyond
			the within-face contact selection described in Sec.~\ref{D2.3.1}. When a new segment
			requires a different face, the robot disengages from the current contact,
			repositions the mobile base, and re-establishes contact on the newly selected
			face before continuing the push. Straight pushing maintains the current face,
			lateral pushing switches to the corresponding side face, and corner pushing
			is realized through successive face transitions. All behaviors share the same
			discrete contact representation and whole-body controller, without requiring
			separately trained pushing policies.

			\begin{figure}[t]
				\centering
				\includegraphics[scale=0.173]{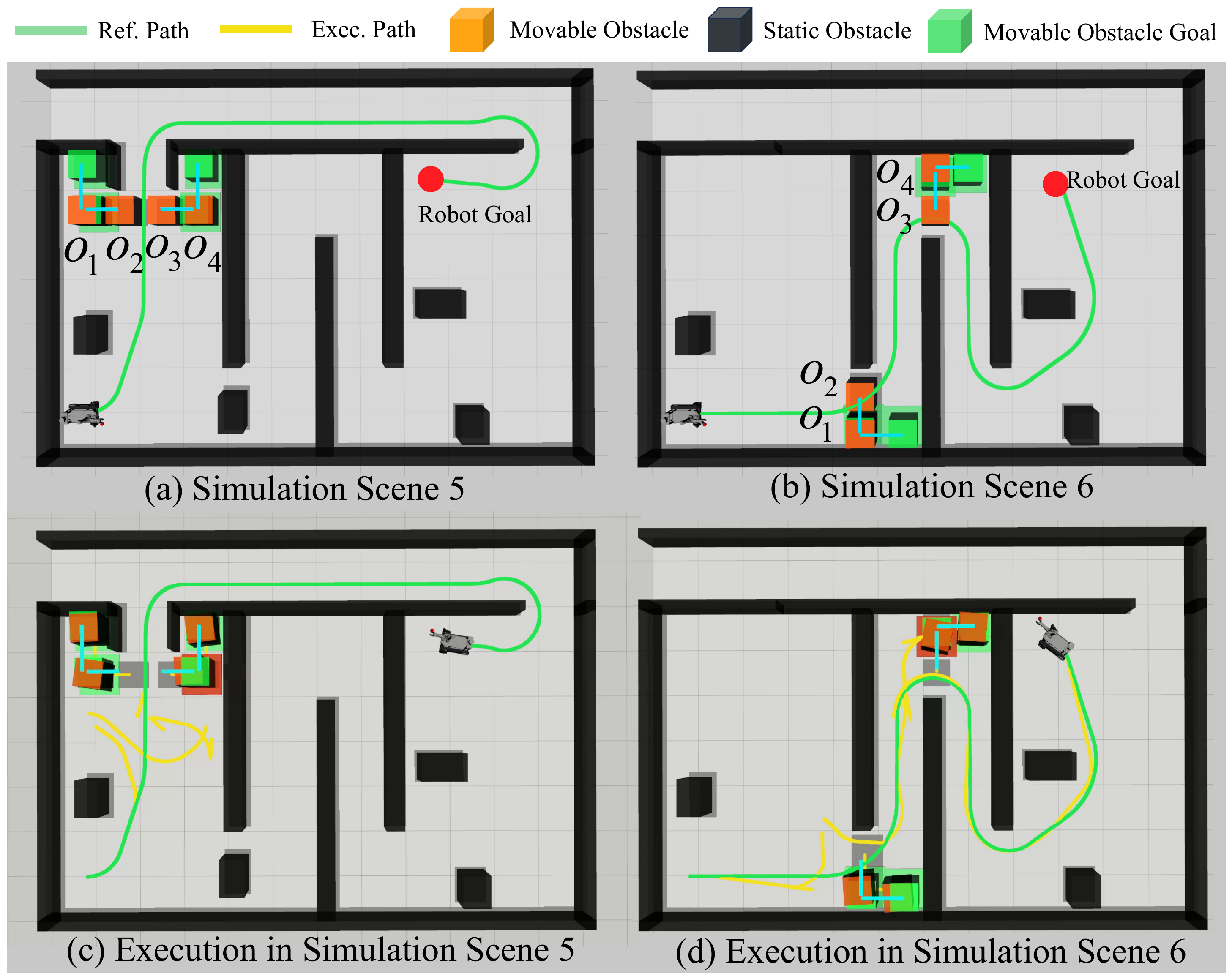}
				\caption{Planning and execution results in dependency-constrained scenarios. (a) and (b) show LLM-assisted planning, where $o_2$ and $o_3$ are key blockers and $o_1$ and $o_4$ are auxiliary obstacles, with $o_1 \prec o_2$ and $o_4 \prec o_3$. (c) and (d) show the simulated execution, in which the planned relocations restore connectivity and enable the robot to reach the goal.}
				\label{f5}
			\end{figure}
			
			\section{Experiment}

			\subsection{Experimental Setup}
			
			We evaluate the proposed NAMO framework in both simulation and real-world
			environments. Simulation experiments and whole-body policy training are conducted in Isaac Gym, with scenes containing fixed obstacles and large movable boxes on known maps to evaluate high-level planning, obstacle relocation, manipulation performance, and integrated task execution. The real-world platform consists of a tracked mobile base and a Unitree
			manipulator. Movable-box poses are estimated online using AprilTag fiducials
			observed by a monocular camera \cite{olsonAprilTagRobustFlexible2011}, while
			the robot pose is provided by a FAST-LIO2-based LiDAR--inertial SLAM system
			\cite{xuFASTLIO2FastDirect2022}. All perception, planning, and control modules
			run onboard an NVIDIA Jetson Orin NX 16GB. We report task success, navigation and manipulation costs, completion time, and motion-related measures where applicable.

			\subsection{High-Level NAMO Planning Evaluation}
			
			
			We first evaluate the proposed high-level NAMO planner in simulation using
			six representative scenarios. Unless otherwise stated, we set
			$\alpha=1.0$, $\beta=0.5$, and $\gamma=1.0$ in
			Eq.~\eqref{eq:placement_cost}, where $\beta$ represents the equivalent cost
			of a contact transition, and $w_n=w_m=1.0$ in
			Eq.~\eqref{eq:task_objective}. Simulation scenes~1--4 contain no obstacle-relocation
			dependencies and evaluate key-blocker identification, manipulation-feasible
			placement search, and the trade-off between detouring and interaction
			(Fig.~\ref{f4}). Simulation scenes~5--6 involve non-local manipulation dependencies and
			evaluate the failure-driven dependency expansion with LLM-assisted reasoning
			(Fig.~\ref{f5}).

			The results in Simulation scenes~1--4 show that the planner selects task-level solutions according to reachability and total execution cost. In simulation scene~1, where no navigation-only path exists, $o_1$ and $o_2$ are identified as critical blockers and feasible relocation plans restore connectivity to the goal. In simulation scene~2, two feasible NAMO alternatives are compared by their combined navigation and manipulation costs, and the lower-cost solution is selected. Simulation scenes~3 and~4 demonstrate the trade-off between detouring and interaction: Simulation scene~3 favors detouring because the reduction in navigation cost is smaller than the additional manipulation cost, whereas the wider corridor in Simulation scene~4 requires only a short relocation of $o_2$, making NAMO more economical. These results confirm that the planner can recover connectivity when interaction is necessary while avoiding unnecessary manipulation when detouring is preferable. Quantitative results are summarized in Table~\ref{tab:high_level_planning}.

			\begin{table}[t]
				\centering
				\caption{Task-cost comparison for high-level planning in Scenes~1--4.}
				\label{tab:high_level_planning}
				\renewcommand{\arraystretch}{0.95}
				\setlength{\tabcolsep}{7pt}
				\begin{tabular}{c l c c c}
					\toprule
					Simulation scene & Strategy
					& $J_{\rm nav}$
					& $J_{\rm man}$
					& $J_{\rm total}$ \\
					\midrule
					1 & NAMO
					& 23.59 & 5.85 & 29.44 \\
					\midrule
					2 & \textbf{NAMO-A}
					& \textbf{15.40} & \textbf{9.67} & \textbf{25.07} \\
					& NAMO-B
					& 23.59 & 8.84 & 32.43 \\
					\midrule
					3 & \textbf{Navigation-only}
					& \textbf{23.59} & \textbf{0} & \textbf{23.59} \\
					& NAMO
					& 15.40 & 9.67 & 25.07 \\
					\midrule
					4 & Navigation-only
					& 23.59 & 0 & 23.59 \\
					& \textbf{NAMO}
					& \textbf{15.06} & \textbf{3.31} & \textbf{18.37} \\
					\bottomrule
				\end{tabular}
			\end{table}
			
			\begin{table}[t]
				\centering
				\caption{Manipulation robustness under varying initial heading offsets and friction conditions.}
				\label{tab:manipulation_robustness}
				\renewcommand{\arraystretch}{0.95}
				\setlength{\tabcolsep}{7pt}
				\begin{tabular}{l c c c}
					\toprule
					Test & Setting
					& Success (\%)
					& $T_{\rm comp}$ (s) \\
					\midrule
					
					Initial heading
					& $-40^\circ$ & 96  & $12.3\pm1.7$ \\
					& $-20^\circ$ & 100 & $9.7\pm1.1$ \\
					& $0^\circ$   & 100 & $5.8\pm0.6$ \\
					& $20^\circ$  & 96  & $10.2\pm0.7$ \\
					& $45^\circ$  & 96  & $13.1\pm1.5$ \\
					\midrule
					
					Friction
					& $\mu=0.3$ & 100 & $6.6\pm0.7$ \\
					& $\mu=0.5$ & 100 & $8.4\pm1.1$ \\
					& $\mu=0.7$ & 96  & $9.9\pm1.4$ \\
					& Nonuniform & 88 & $13.3\pm3.8$ \\
					
					\bottomrule
				\end{tabular}
			\end{table}

			Simulation scenes~5--6 further evaluate dependency-constrained relocation. As shown in Fig.~\ref{f5}, in Simulation scene~5, although $o_2$ and $o_3$ are the direct blockers, their feasible placement and manipulation spaces are occupied by $o_1$ and $o_4$, causing geometry-only planning to fail. With LLM-assisted dependency reasoning, the planner identifies $o_1$ and $o_4$ as prerequisite relocations and successfully recovers a feasible solution. The same outcome is observed in Simulation scene~6. These results show that LLM-assisted reasoning enables the planner to uncover non-local manipulation dependencies that cannot be resolved by local geometric search alone, while all resulting relocation plans remain verified by deterministic planning.

			\begin{figure}[t]
				\centering
				\includegraphics[scale=0.21]{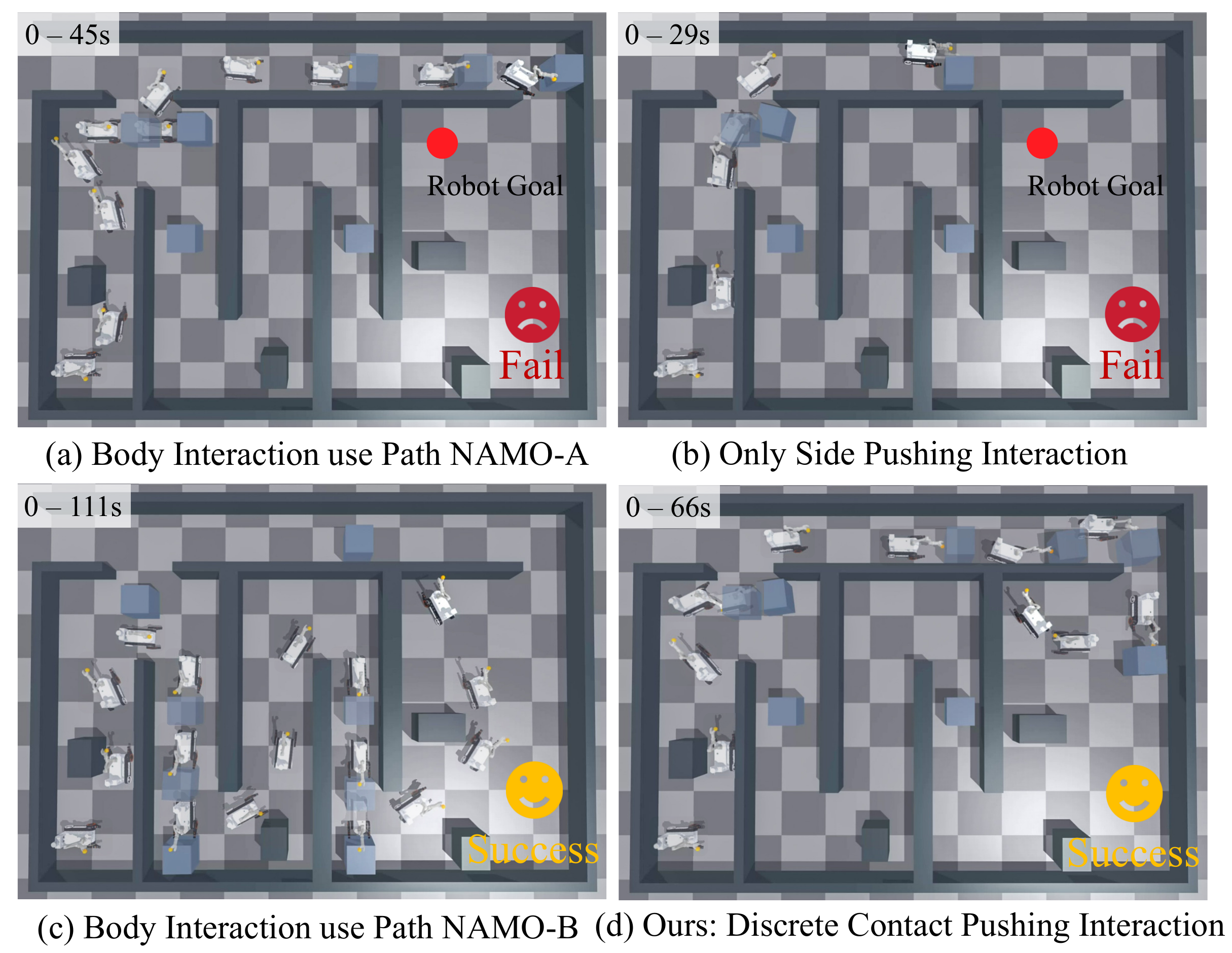}
				\caption{Interactive navigation comparison in Simulation Scene~2. (a) and (c) Robot-body pushing, (b) arm-based side pushing~\cite{biInteractiveNavigationLegged}, and (d) our discrete-contact pushing. Panels (a), (b), and (d) execute NAMO-A, while (c) executes NAMO-B.}
				\label{f8}
			\end{figure}
			
			\begin{figure*}[t]
				\setlength{\belowcaptionskip}{-0.5cm}
				\vspace{0.2cm}
				\centering
				\includegraphics[scale=0.35]{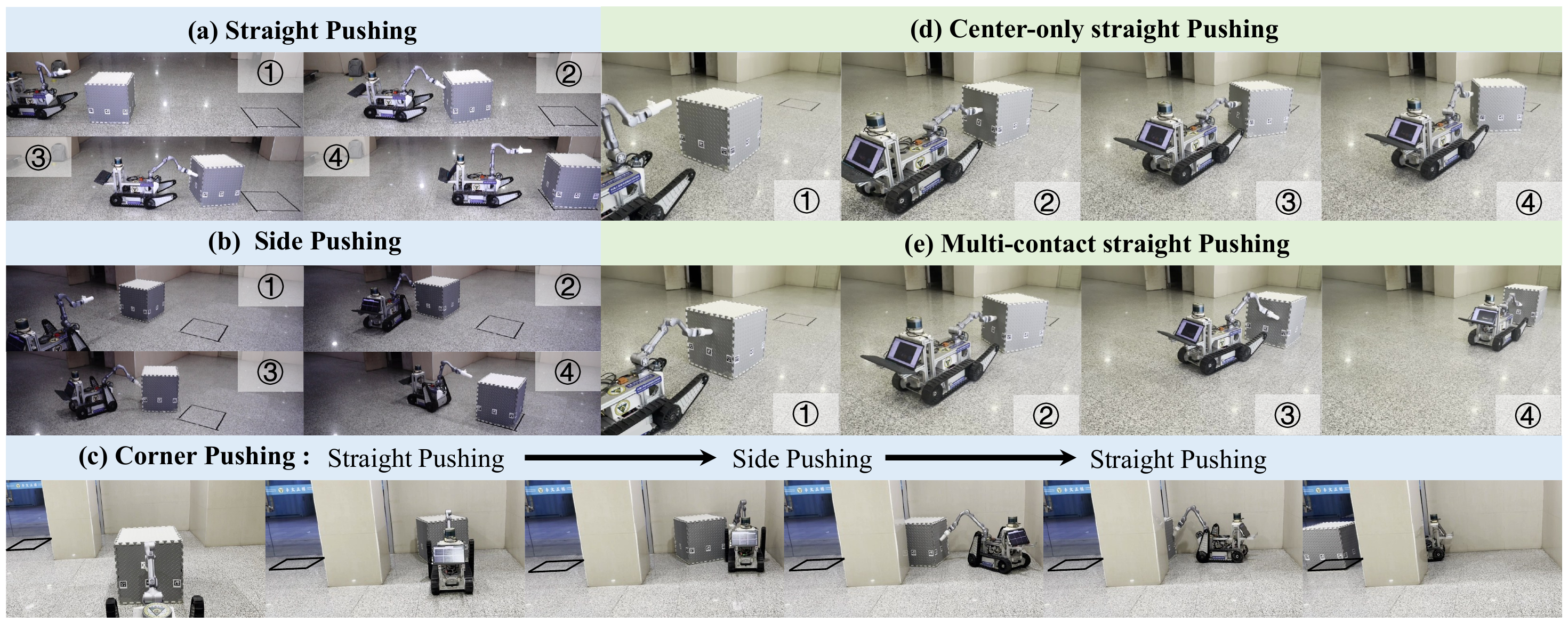}
				\caption{Real-robot obstacle-pushing skills. (a) and (b) demonstrate straight and side pushing, respectively. (c) shows corner pushing, where an intermediate relocation pose enables successive straight, side, and straight pushing to guide the obstacle through the corner. (d) and (e) compare straight pushing with center-only and LCR contact, respectively.}
				\label{f6}
			\end{figure*}
			
			\begin{figure*}[t]
				\setlength{\belowcaptionskip}{-0.5cm}
				\vspace{0.2cm}
				\centering
				\includegraphics[scale=0.4]{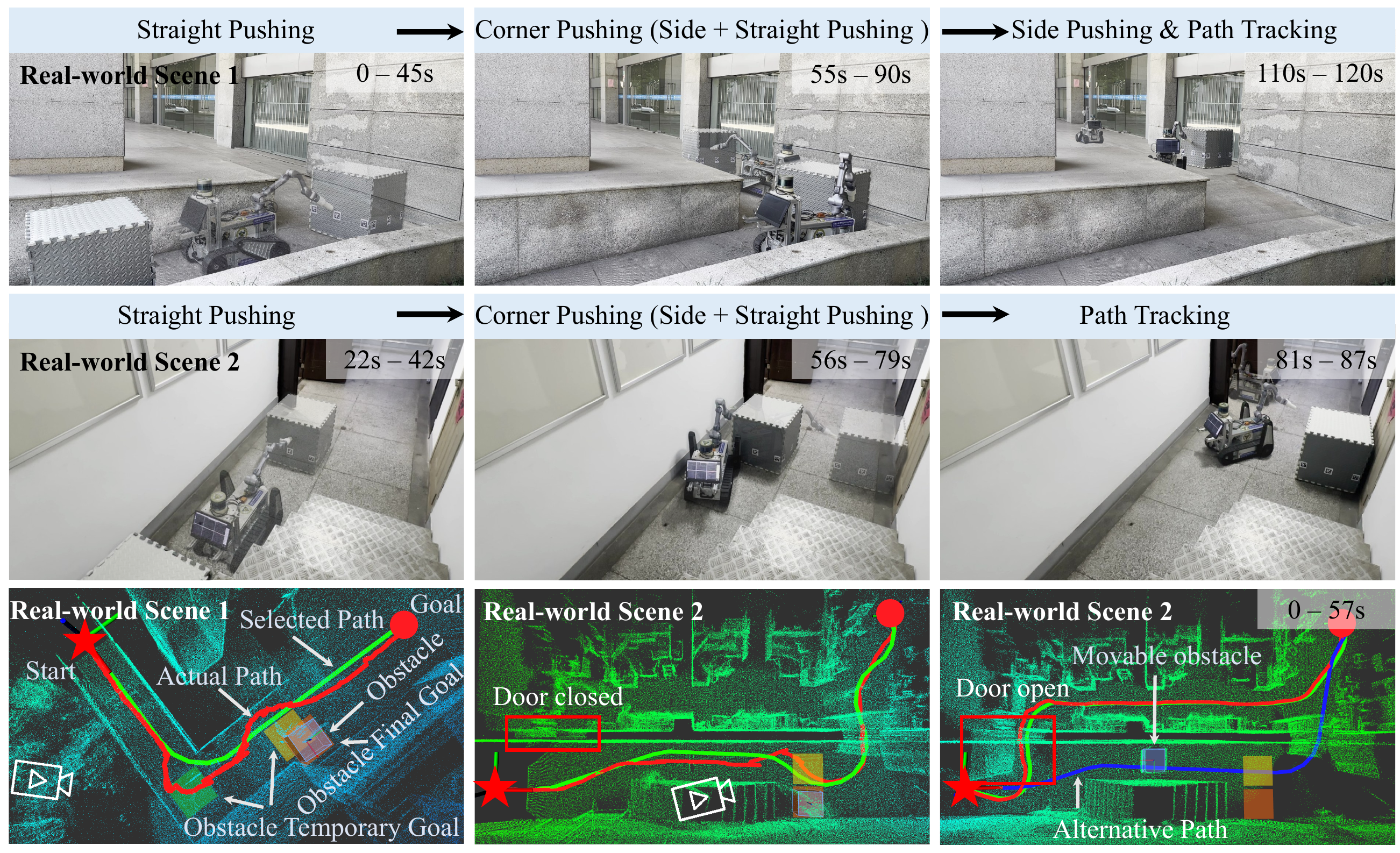}
				\caption{Real-world integrated NAMO execution. Scene~1 and the closed-door configuration of Scene~2 contain no feasible navigation-only route; the robot sequentially relocates obstacles through two and one intermediate relocation poses, respectively, to restore connectivity and reach the goal. In the open-door configuration of Scene~2, opening the door creates a lower-cost navigation-only route, which is selected by the planner.}
				\label{f7}
			\end{figure*}
			
			\subsection{Interactive Manipulation Evaluation}

			We first evaluate the interaction capabilities enabled by the proposed discrete contact modes. As shown in Fig.~\ref{f6}(a)--(c), straight pushing drives the box along the primary motion direction while correcting orientation deviations when necessary. Side pushing relocates the box transversely by switching to the corresponding side face, whereas corner pushing changes the motion direction through contact disengagement, base repositioning, and re-contact. The importance of contact-location flexibility is further examined in Fig.~\ref{f6}(d) and (e). Center-only pushing cannot recover from accumulated heading deviations, whereas LCR-contact adjusts among the left, center, and right anchors to regulate the object pose and successfully reach the target. This confirms that the improved maneuverability arises from the additional contact-location freedom rather than a different controller.

			Robustness is further evaluated in simulation under varying initial heading offsets and friction conditions, with 25 trials per setting and a $15\,\mathrm{s}$ timeout for failure. As shown in Table~\ref{tab:manipulation_robustness}, success remains at $96$--$100\%$ across all tested heading offsets and above $96\%$ under uniform friction variations. Under nonuniform friction, success decreases to $88\%$ and completion time increases to $13.3\pm3.8\,\mathrm{s}$, mainly due to repeated corrections induced by asymmetric object motion.

			Finally, we examine whether these manipulation capabilities translate into more effective NAMO execution in Simulation Scene~2. As shown in Fig.~\ref{f8}, under the same relocation poses and NAMO plans from Sec.~\ref{M2.2.2}, robot-body pushing and arm-based side pushing~\cite{biInteractiveNavigationLegged} both fail on NAMO-A because they cannot complete the compound relocation of $o_2$ in the narrow corridor. Robot-body pushing succeeds only with the longer NAMO-B solution, requiring $111\,\mathrm{s}$, whereas our discrete-contact method directly executes NAMO-A in $66\,\mathrm{s}$. This demonstrates that manipulation versatility is essential for converting efficient high-level NAMO plans into executable navigation.

			\subsection{Real-World Integrated NAMO Evaluation}

			Finally, we evaluate the complete NAMO system in two real-world L-shaped environments, as shown in Fig.~\ref{f7}. In Real-world Scene~1, the corridor is fully blocked and no navigation-only path exists. The planner therefore selects obstacle relocation, and the robot clears the passage by combining forward pushing, contact disengagement, base repositioning, and lateral re-contact before replanning to the goal, demonstrating consistent execution of the planned NAMO behavior.
			
			Real-world scene~2 further evaluates the navigation--interaction trade-off under two door configurations. With the door open, the navigation-only and NAMO paths have similar lengths, so direct navigation is selected because manipulation would introduce additional cost. With the door closed, navigation-only becomes infeasible; the robot instead clears the passage through a short forward push and lateral adjustment, then replans to the goal. Together, these experiments show that the proposed framework can adapt its task-level decision to environmental reachability and execution cost while maintaining consistency between high-level planning and physical execution.

			\section{CONCLUSIONS}
			
			We presented a hierarchical NAMO framework for mobile manipulators operating among large, non-graspable box-shaped obstacles. The framework couples manipulation-feasible high-level planning with discrete-contact whole-body pushing, while selectively using an LLM to expand non-local manipulation dependencies when direct relocation fails. Simulation results demonstrate effective trade-offs between navigation and interaction, successful resolution of dependency-constrained cases, and improved pose regulation through multi-contact pushing. Real-world experiments further validate the consistency between task-level planning, obstacle relocation, and physical execution. Future work will investigate more adaptive manipulation under uncertain contact dynamics and extend the framework to objects with more diverse geometries and less structured environments.

		\end{document}